\pdfoutput=1
\documentclass[11pt]{article}

\usepackage{acl}

\usepackage{times}
\usepackage{latexsym}

\usepackage[T1]{fontenc}
\usepackage[utf8]{inputenc}

\usepackage{microtype}

\usepackage{bm}
\usepackage{comment}
\usepackage{dingbat}
\usepackage{booktabs}
\usepackage{graphicx}
\usepackage{multirow}
\usepackage{amsfonts}
\usepackage{mathtools}
\usepackage[normalem]{ulem}
\usepackage[USenglish]{babel}
\usepackage[export]{adjustbox}
\usepackage[justification=centering]{caption}
\usepackage{stfloats}

\usepackage{pifont}
\newcommand{\cmark}{\ding{51}}
\newcommand{\xmark}{\ding{55}}

\title{\centering Speech Translation with Foundation Models and Optimal Transport: \\ UPC at IWSLT23}

\author{\phantom{-----------------------} Ioannis Tsiamas, Gerard I. Gállego, José A. R. Fonollosa \\
    \phantom{-----------------------} Universitat Politècnica de Catalunya, Barcelona \\
	\phantom{-----------------------} \small\texttt{\{ioannis.tsiamas,gerard.ion.gallego,jose.fonollosa\}@upc.edu}
    \\\And \hspace{5cm} Marta R. Costa-jussà \\
    \hspace{5cm} Meta AI, Paris \\
	\hspace{5cm} \small\texttt{costajussa@meta.com}
 }

\begin{document}
\maketitle
\begin{abstract}
    This paper describes the submission of the UPC Machine Translation group to the IWSLT 2023 Offline Speech Translation task. Our Speech Translation systems utilize foundation models for speech (wav2vec 2.0) and text (mBART50). We incorporate a Siamese pretraining step of the speech and text encoders with CTC and Optimal Transport, to adapt the speech representations to the space of the text model, thus maximizing transfer learning from MT. After this pretraining, we fine-tune our system end-to-end on ST, with Cross Entropy and Knowledge Distillation. Apart from the available ST corpora, we create synthetic data with SegAugment to better adapt our models to the custom segmentations of the IWSLT test sets. Our best single model obtains 31.2 BLEU points on MuST-C tst-COMMON, 29.8 points on IWLST.tst2020 and 33.4 points on the newly released IWSLT.ACLdev2023.
\end{abstract}

\section{Introduction}
\label{sec:intro}
    
    In the past decade, the field of Speech Translation (ST) has seen significant advancements, mainly due to end-to-end models that directly translate speech, offering a more efficient method compared to traditional cascade systems \cite{taking_stock}. Despite data availability challenges, recent progress has diminished the performance disparity between these approaches \cite{cascade,srpol2020,espnet-2021,iwslt2020}. Critical to the advancements in end-to-end models is the exploitation of ASR and MT data through pretraining strategies \cite{asr-pretrain,asr-pretrain-ex1,asr-pretrain-ex2,s-transformer,fairseq-s2t,afs,no-mt-pretrain}.
    
    Recently, \citet{ctc-ot} proposed a method to effectively utilize both ASR and MT pretraining to enhance ST. This approach involves pretraining an encoder-decoder MT system with available text data, followed by pretraining a speech encoder to generate representations similar to the MT system's encoder (\textit{Siamese pretraining}) using Connectionist Temporal Classification (CTC) supervision \cite{ctc} and Optimal Transport \cite{ot}. The resulting speech encoder and text decoder can be fine-tuned with ST data.
    
    Another way of incorporating ASR and MT is to leverage large pretrained speech and text models as a foundation for end-to-end ST systems \cite{lna, upc2021, chimera, yitrans2022, kit2022, upc2022}. However, these systems encounter representation discrepancy issues, which can hinder the full exploitation of pretrained foundation models. \citet{upc2021,m-adapter} aimed to address this by adding \textit{coupling modules} after the pretrained encoder, while other focus on solving the length discrepancies \cite{afs, sate, gaido-etal-2021-ctc}. \citet{chimera} tackled the issue by projecting speech and text features to a common semantic space using attention mechanisms and semantic memories.
    
    \begin{figure*}[ht]
        \centering
        \includegraphics[width=\textwidth]{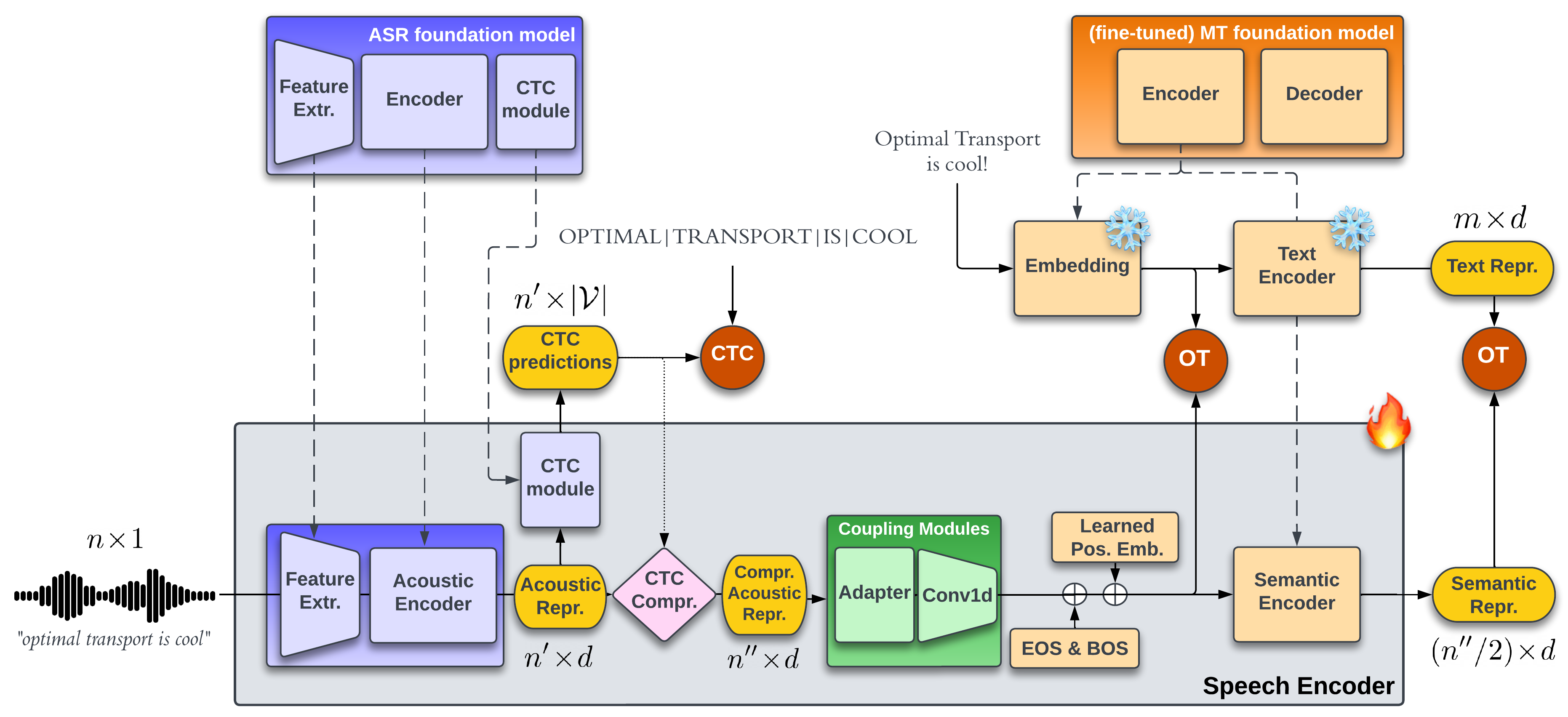}
        \caption{Extended Siamese pretraining}
        \label{fig:ot}
    \end{figure*}
    
    In our work, we tackle the issue of misaligned speech and text encoder representations by adopting the approach proposed by \citet{ctc-ot}. Our system uses a speech foundation model fine-tuned on English ASR, wav2vec 2.0 \cite{wav2vec2.0}, and an MT foundation model fine-tuned on multilingual MT (En-Xx), mBART50 \cite{mbart50}, as described in Section \ref{subsec:architecture}. Building on prior research \cite{sate,chimera}, we employ two encoders: an acoustic encoder from wav2vec 2.0 and a semantic encoder from mBART50. Coupling modules link these encoders to address length discrepancy. We extend \citet{ctc-ot} by applying CTC and OT losses to the outputs of the acoustic and semantic encoders, respectively, add a second auxiliary OT loss for the inputs of the semantic encoder, and keep the text encoder frozen to keep the MT space intact. This method aligns the speech encoder's representations with the MT foundation model, effectively improving the final ST system's performance by mitigating representation mismatch.
    
    In summary, we participate in the IWSLT 2023 Offline Speech Translation task, focusing on translating spoken English to written German, by employing an end-to-end system. We leverage ASR and MT foundation models with the Siamese pretraining approach, to effectively bring their encoder's representations closer. We furthermore decouple acoustic and semantic modeling in our speech encoder, adjust for the length miss-match between speech and text with several coupling modules, and apply \textit{knowledge distillation} \cite{kd_original} from MT \cite{kd2019, kd2020}, using mBART50.

\section{Methodology}
    \label{sec:methodology}

    \begin{figure}[t]
        \centering
        \includegraphics[width=\columnwidth]{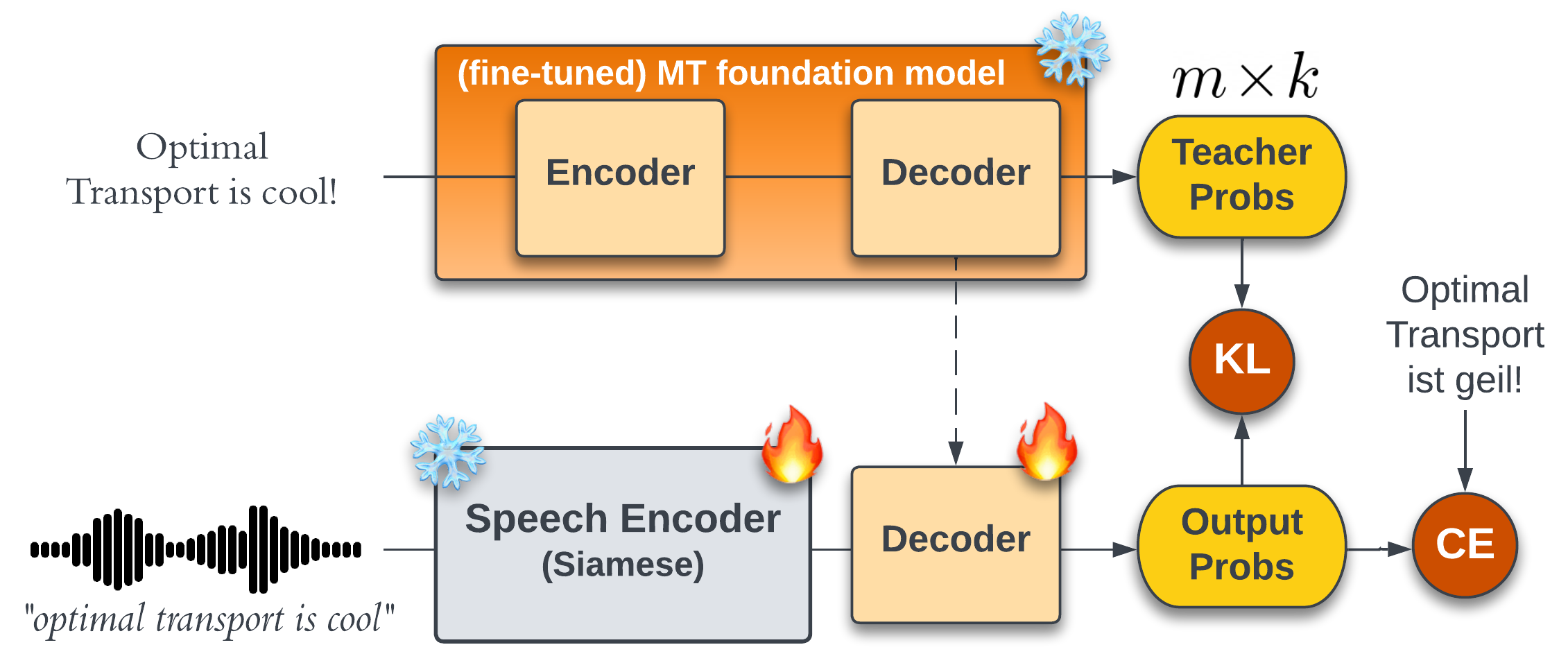}
        \caption{Speech Translation fine-tuning}
        \label{fig:st}
    \end{figure}

    %Our system is an encoder-decoder transformer architecture, leveraging ASR and MT foundation models (\S \ref{subsec:architecture}). We first train the speech encoder with the Extended Siamese pretraining method (\S \ref{subsec:ot}), and then fine-tune it with the MT decoder on end-to-end ST (\S \ref{subsec:ft}).
    Our system, an encoder-decoder transformer, leverages ASR and MT foundation models (\S \ref{subsec:architecture}). We initially train the speech encoder with an Extended Siamese pretraining (\S \ref{subsec:ot}), and then fine-tune it with the MT decoder for end-to-end ST (\S \ref{subsec:ft}).
    
    \subsection{System architecture}
    \label{subsec:architecture}

        As depicted in Figures \ref{fig:ot} and \ref{fig:st}, the encoder of our system is composed of several interconnected modules, while the decoder is adopted directly from the MT foundation model. The speech encoder is designed to generate representations closely resembling those of the MT foundation model, ensuring better compatibility between them. The following paragraphs provide a detailed overview of its key components and their functions.

        \paragraph{Acoustic Modeling} The speech waveform $x \in \mathbb{R}^{n}$ is first processed by a feature extractor, which consists of several strided convolutional layers, downsampling the input to a length of $n'$. Following, a Transformer encoder with dimensionality $d$ is responsible for the acoustic modeling. Both these modules are initialized from an ASR foundation model.

        \paragraph{CTC Compression} The obtained acoustic representation $h \in \mathbb{R}^{n' \times d}$ is passed through a linear layer (initialized from the ASR model) and a softmax to generate the ASR vocabulary predictions $p^{(ctc)} \in \mathbb{R}^{n' \times |\mathcal{V}|}$, where $\mathcal{V}$ is the size of the vocabulary. We apply CTC compression \cite{gaido-etal-2021-ctc} to the acoustic representation, averaging the representations corresponding to repeating predictions on $p^{(ctc)}$ and removing those associated with the blank token. This process results in a new compressed representation $h^{(compr)} \in \mathbb{R}^{n'' \times d}$, where $n''$ denotes the compressed length of the sequence. This compression helps to reduce the length discrepancy between speech and text representations, which, in turn, facilitates the alignment process during Siamese pretraining (\S \ref{subsec:ot}).
        
        \paragraph{Coupling Modules} Next, we apply an \textit{adapter} \cite{adapters}, consisting of a linear projection to $8d$, a non-linear activation, a linear projection back to $d$. This module serves to (1) process the collapsed representations resulting from the compression and (2) provide sufficient parameters between the CTC and first OT loss to decouple their influence (\S \ref{subsec:ot}). After the adapter we apply a strided 1D Convolution that subsamples the sequence by a factor of 2, which can help transform it closer to a sub-word level representation, rather than a character-level one, and subsequently aid in the Optimal Transport training with the sub-word level representation from the text encoder (\S \ref{subsec:ot}).
        
        \paragraph{Semantic Modeling} At this point, we modify the representation to better match the input expected by the MT encoder. This is achieved by prepending and appending special tokens that correspond to the BOS and EOS tokens used in MT. We also re-introduce positional information to the representation with learned positional embeddings. Both the special tokens $t^{bos}, t^{eos} \in \mathbb{R}^d$ and the positional embeddings $E^{pos} \in \mathbb{R}^{(M+2) \times d}$ (with $M$ representing the maximum sequence length) are learnable parameters initialized from the MT foundation model. The motivation is to bring the representation closer to the text embedding from the MT model, facilitating OT loss convergence (\S \ref{subsec:ot}). Finally, the representation is processed by several more transformer encoder layers, which are initialized from the MT model and are responsible for semantic modeling.

    \subsection{Siamese pretraining}
    \label{subsec:ot}

        Our approach builds upon the Siamese pretraining proposed by \citet{ctc-ot}, which exploits both ASR and MT pretraining to improve ST performance. This approach involves pretraining the encoder of an ST system jointly with Connectionist Temporal Classification (CTC) and Optimal Transport (OT), bringing its representations close to those of an MT encoder. This pretraining strategy has demonstrated superior results compared to traditional ASR pretraining with encoder-decoder and Cross-Entropy \cite{ctc-ot}. In this work, we build upon the method of \citet{ctc-ot} in several ways. First, we decouple the CTC and OT losses to correspond to the acoustic and semantic representations. Second, we add an extra auxiliary OT loss to better adapt the input to the semantic encoder. Next, we also employ CTC-based compression and coupling modules to better align the length of speech features with corresponding sub-word text representations. Finally, we opt to freeze the text encoder to not modify the MT decoder's representation space. The extended Siamese pretraining scheme is illustrated in Figure \ref{fig:ot}. For brevity, we refer to it simply as "Siamese" throughout the rest of the paper.

        The Siamese pretraining is supervised by a combination of loss functions, each serving a distinct purpose. The CTC loss ensures the performance of the acoustic modeling by applying to the predictions of the CTC module. Meanwhile, the two OT losses target the input and output of the semantic encoder, and aim to align them with the text encoder representations. We calculate the OT loss as the Wasserstein distance \cite{wasserstein} between the text and speech representations, using an upper bound approximation, which is efficiently evaluated by the Sinkhorn algorithm \cite{sinkhorn}. Since the Wasserstein distance is position invariant, we follow \cite{ctc-ot}, and apply positional encodings, to make it applicable to sequences. The combined loss function for the Siamese pretraining stage is given by:

        \begin{align} \label{eq:ot_loss}
            \mathcal{L}^{siamese} &= \alpha \, \mathcal{L}^{CTC} + \beta \, \mathcal{L}^{OT_1} + \gamma \, \mathcal{L}^{OT_2}
        \end{align}

        Where $\alpha$, $\beta$, and $\gamma$ are hyperparameters that control the relative importance of each loss component in the combined pretraining loss.

    \subsection{Speech Translation fine-tuning}
    \label{subsec:ft}

        Upon obtaining the encoder from \S \ref{subsec:ot}, we utilize it to initialize our ST system's encoder, while using the MT foundation model to initialize the decoder (Fig. \ref{fig:st}). In addition to the Cross Entropy loss, we optionally provide guidance for the ST training through Knowledge Distillation (KD) \cite{topk}, using the MT foundation model as a teacher. Specifically, we only use the top-$k$ predictions rather than the entire distribution, and soften them using a temperature $T$ \cite{kd2020}.
        
        Since CTC supervision is not employed at this stage, we freeze the Feature Extractor, Acoustic Encoder, and CTC module from our encoder. During training, we optimize the parameters of the ST system's encoder and decoder with respect to the combined loss function, which is the sum of the Cross Entropy loss and the optional KD loss:

        \begin{align} \label{eq:st_loss}
            \mathcal{L}^{ST} &= \lambda \, \mathcal{L}^{CE} + (1 - \lambda) \, \mathcal{L}^{KL}
        \end{align}

        Where $\mathcal{L}^{CE}$ is the Cross Entropy loss, $\mathcal{L}^{KL}$ is the Kullback–Leibler divergence between the MT and ST output distributions, and $0 \leq \lambda \leq 1$ is a hyperparameter that controls the relative importance of each loss component in the combined ST loss.

\section{Data}
\label{subsec:data}
    \subsection{Datasets}
    \label{subsec:datasets}
            
        To train our ST models we used data from three speech translation datasets, MuST-C v3 \cite{mustc}, Europarl-ST \cite{europarl} and CoVoST-2 \cite{covost}. MuST-C is based on TED talks, Europarl-ST on the European Parliament proceedings, and CoVoST is derived from the Common Voice dataset \cite{commonvoice}. Their statistics are available in the first part of Table \ref{tbl:train_data}. We use as development data the IWSLT test sets of 2019 and 2020 \cite{iwslt2019,iwslt2020}, which are based on TED talks, and the ACL development set of 2023, which contains 5 presentations from ACL 2022. All development data are unsegmented, meaning that they are long and continuous speeches. We apply SHAS segmentation (\S \ref{sec:segmentation}) before translating them. For the Siamese pretraining, we used the English ASR data from MuST-C v3 and Europarl-ST, as well as CommonVoice v11 \cite{commonvoice} (Table \ref{tbl:train_data}).

    \subsection{Data Augmentation} \label{subsec:data_aug}

        We employ data augmentation, to create more ST data for training our models (Table \ref{tbl:train_data}). We use the MT foundation model, to translate the transcript of English CommonVoice v11 \cite{commonvoice}. Since CommonVoice data contains various accents, we expect the synthetic data will be helpful for translating the ACL talks domain, which has predominantly non-native English accents. We additionally utilize SegAugment \cite{segaugment}, which creates alternative versions of the training data by segmenting them differently with SHAS \cite{shas}. We apply SegAugment to MuST-C v3, with three different length parameterizations: \textit{medium (m)} (3 to 10 seconds), \textit{long (l)} (10 to 20 seconds), and \textit{extra-long (xl)} (20 to 30 seconds). We expect that SegAugment will be beneficial for translating the SHAS-segmented test sets, due to the similar segmentations of the training data it provides, as shown in \citet{segaugment}.

        \begin{table}[h]
            \centering
            \resizebox{0.9\columnwidth}{!}{
            \begin{tabular}{lccc}
            \toprule
                                                & \textbf{Original}             & \textbf{Siamese}                  & \textbf{ST}                   \\
            \midrule
            \textbf{ST datasets}                &                               &                                   &                               \\
            MuST-C v3                           & $427$                         & $417$                             & $421$                         \\
            $\hookrightarrow$ SegAugment        & $1,364^{\dagger}$             & $-$                               & $1,007^{\dagger}$             \\
            Europarl-ST                         & $77$                          & $64$                              & $75$                          \\
            CoVoST 2                            & $362$                         & $-$                               & $344$                         \\
            \midrule
            \textbf{ASR datasets}               &                               &                                   &                               \\
            CommonVoice v11                     & $1,503$                       & $1,361$                           & $1,082^{\dagger}$             \\
            \midrule
            \textbf{Total}                      & $-$                           & $1,842$                           & $2,929$                       \\
            \bottomrule
            \end{tabular}
            }
            \caption{Filtered training data (in hours) for Siamese and ST training stages. Synthetic data is denoted with $\dagger$.}
            \label{tbl:train_data}
        \end{table}

    \subsection{Data Filtering}
    \label{subsec:filtering}
        
        \paragraph{Siamese pretraining} We remove speaker names, as well as events like "Laughter" and "Applause", we convert numbers to their spelled-out forms,\footnote{\url{https://github.com/savoirfairelinux/num2words}} convert all text to lowercase, and finally remove all characters that are not included in the vocabulary of the ASR foundation model. Furthermore, we apply a step of ASR-based filtering, to filter out noisy examples stemming from wrong audio-text alignments, where we remove examples with high word-error-rate (WER). We adjust the threshold for each dataset dynamically, ensuring that the resulting data has a WER of 0.11. Thus, the thresholds are 0.5 for MuST-C, 0.28 for Europarl-ST, and 0.4 for CommonVoice, which indicates that Europarl-ST has a significant number of misalignments, a conclusion supported by manual inspection. Removing them allowed for faster convergence during Siamese pretraining.
        
        \paragraph{ST fine-tuning} We apply text normalization to the original ST data, remove speaker names and event-related tags from the MuST-C dataset, discard examples with extreme source-to-target text length ratios \cite{fbk22}, and finally remove audio-transcription misaligned examples with ASR-based filtering, using a fixed WER threshold of 0.5. For the synthetic CommonVoice data, we remove the ones already present in CoVoST. We also filter the synthetic examples of SegAugment, as the SHAS segmentation frequently resembles the original segmentation, thus resulting in highly similar examples. We retain only the ones that are sufficiently dissimilar from the original ones, based on text similarity measures, using TF-IDF features from the translations. More concretely, for each talk id, we compute the similarity matrix of its original translations and the new candidates from SegAugment, find the most similar original example for each new candidate, and add it to the filtered data only if its similarity score is below 0.8. We apply this approach also between the different SegAugment versions (\textit{m, l, xl}).

\section{Experiments} \label{subsec:experiments}
        
    Here we describe the experiments we carried out in this work. The implementation details are available in \S \ref{subsec:impl_details}.

    \paragraph{IWSLT '22 System} For the IWSLT 2022 offline task, our submission employed a HuBERT encoder \cite{hubert} and an mBART50 (En-Xx) decoder, which were efficiently fine-tuned to ST with the LNA strategy \cite{lna} and parallel adapters \cite{adapters_parallel}, using datasets such as MuST-C v2, Europarl-ST and CoVoST. The architecture included three 1D convolutional layers between the encoder and decoder, resulting in a subsampling of the encoder representation by a factor of 8. The final ensemble also comprised models utilizing Knowledge Distillation and a wav2vec 2.0 encoder \cite{upc2022}.  

    \paragraph{Baseline} Our baseline has four main differences compared our last year's best system. We did an initial exploratory analysis of various encoders (\S \ref{sec:app_preliminary}), including different versions of wav2vec 2.0, and HuBERT. Upon observing no significant differences, we opted to utilize wav2vec 2.0 fine-tuned with pseudo-labels \cite{wav2vec-selftraining}, a more prevalent choice within the research community. Despite the strong performance demonstrated by efficient fine-tuning with LNA and parallel adapters, we chose to switch to standard ST fine-tuning in order to optimize performance. Moreover, we employ a semantic encoder initialized from the MT model. Lastly, we also pre-train the foundation models, wav2vec 2.0 with CTC on the ASR data of MuST-C, and mBART50 on the parallel text of MuST-C. It is important to note that only MuST-C data was utilized for the baseline.

    \paragraph{Siamese Pre-training} Instead of pre-training the speech encoder with CTC only, we follow the Siamese pre-training method (\S \ref{subsec:ot}), with the encoder architecture described in \S \ref{subsec:architecture}, to align the encoder representations with the MT model's representation space. The system, instead of using three layers of 1D convolutions, now incorporates also CTC-based compression, a large adapter, and finally a single layer of 1D convolutions. Following the Siamese pre-training on MuST-C's ASR data, we jointly fine-tune the model and the MT decoder on the MuST-C ST data. Similar to the baseline, the MT model is also fine-tuned on the parallel text of MuST-C beforehand.

    \paragraph{More Data} We extend the previously described process by incorporating additional data. Initially, we fine-tune mBART50 using all the MT data (Table \ref{tbl:mt_data}). Subsequently, we perform the Siamese pre-training and ST fine-tuning employing all the available speech data (Table \ref{tbl:train_data}). By incorporating a larger dataset, we aim to enhance the system's generalization capabilities and overall performance.

    \paragraph{Data Augmentation} We employ two data augmentation techniques to increase the performance of our system during ST fine-tuning (\S \ref{subsec:data_aug}), while no modifications are made to the Siamese pre-training. First, we investigate the use of SegAugment \cite{segaugment}, which we apply to MuST-C v3. Secondly, we generate synthetic data from Common Voice \cite{commonvoice}, by leveraging the fine-tuned mBART50 (\S \ref{sec:app_mt}).
    
    \paragraph{KD} We use knowledge distillation with the fine-tuned mBART50 as the teacher (\S \ref{sec:app_mt}). The loss for training the ST model is the average of the standard cross entropy and the Kullback-Leibler (KL) divergence between the MT and ST output probability distributions. We utilize all available ST data in this experiment, including both real and synthetic data.

\section{Audio Segmentation} \label{sec:segmentation}

\begin{figure}[t]
        \centering
        \includegraphics[width=\columnwidth]{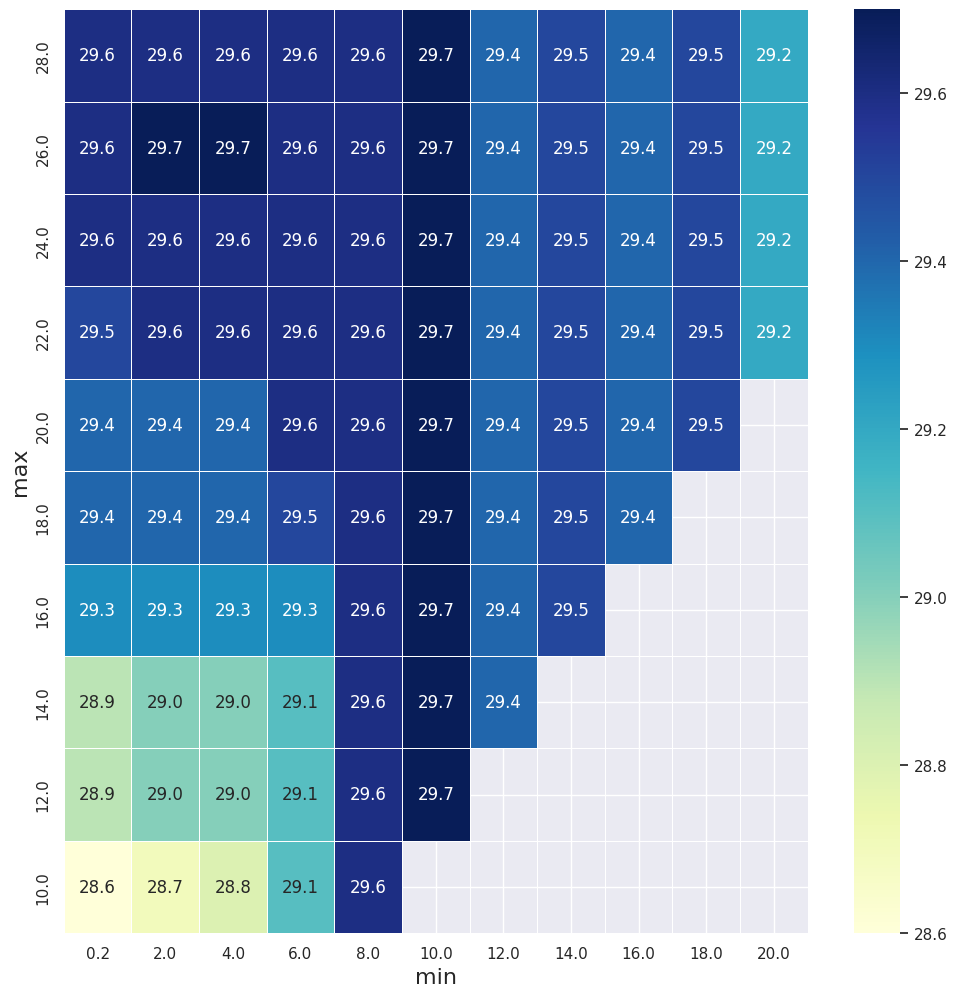}
        \caption{BLEU scores on IWSLT.tst2020 for different combinations of min and max segment length parameters of SHAS.}
        \label{fig:shas_tst2020}
    \end{figure}

\begin{figure}[t]
        \centering
        \includegraphics[width=0.8\columnwidth]{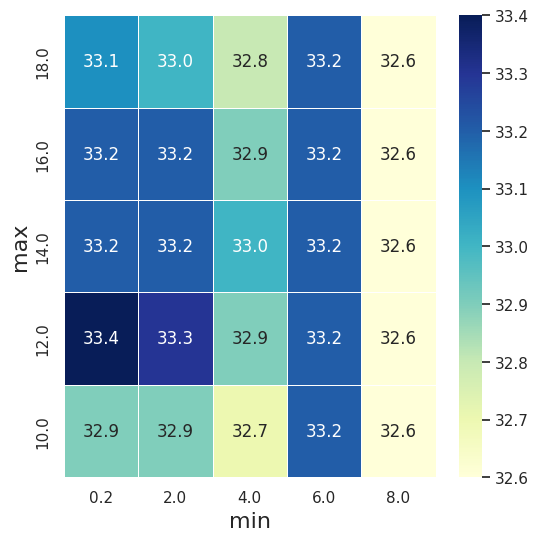}
        \caption{BLEU scores on IWSLT.ACLdev2023 for different combinations of min and max segment length parameters of SHAS.}
        \label{fig:shas_acl2023}
    \end{figure}

    To segment the audio of the IWSLT test sets, we use SHAS \cite{shas}. The tst2023 test set, unlike previous years, contains another two domains apart from TED talks, which are ACL presentations and Press conferences. We tune the parameters of SHAS separately for each domain, but since no development set is available for the press conferences, we decided to treat it as the ACL domain. For fine-tuning the segmentation parameters, we used the ST model that was trained with synthetic data from CommonVoice and SegAugment and initialized from Siamese pre-training (Table \ref{tab:results}, 2d). We evaluate the performance of the ST model on many different combinations of the min and max segment length parameters, between 0.2-30 seconds on IWSLT.tst2019 and 0.2-18 on ACLdev2023. In Figure \ref{fig:shas_tst2020}, we observe that the minimum segment length of 10 seconds is consistently reaching the best BLEU of 29.7 points. We decided to choose the combination of 10-26 seconds, since the max of 26, seemed to be slightly better compared to other neighboring values. As depicted in Figure \ref{fig:shas_acl2023}, smaller segments are better for the ACL domain, with the best BLEU score obtained for min of 0.2 and max of 12. We hypothesize that the differences in the optimal segmentation between the IWSLT and ACL sets is because the ACL data are essentially out-of-domain for our ST models. In turn, the ST models are not confident in their predictions to handle long segments, and thus it is better to translate short segments instead.

\section{Results} \label{sec:results}

    \begin{table*}[t]
        \centering
        \resizebox{0.75\textwidth}{!}{
        \begin{tabular}{@{}lllccccc@{}}
        \toprule
                           & \textbf{} & \textbf{\textbf{Dataset}}            & \multicolumn{2}{c}{\textbf{MuST-C}} & \multicolumn{3}{c}{\textbf{IWSLT}}                                          \\ \midrule
                           &           & \textit{split}                       & v2               & v3               & tst2019                 & tst2020                 & acl2023                 \\ \midrule
        \multicolumn{3}{l}{\textbf{UPC '22} \cite{upc2022}}                   &                  &                  &                         &                         &                         \\
        \multirow{2}{*}{0} & a         & Best Single                          & $29.4$           & -                & $24.9$                  & $26.8$                  & -                       \\
                           & b         & Best Ensemble                        & $30.8$           & -                & $25.4$                  & $27.8$                  & -                       \\ \midrule
        \multicolumn{3}{l}{\textbf{Only MuST-C}}                              &                  &                  &                         &                         &                         \\
        \multirow{2}{*}{1} & a         & Baseline                             & $29.8$           & $29.9$           & $25.7$                  & $27.3$                  & $25.1$                  \\
                           & b         & 1a + Siamese Pretraining & $30.8$           & $30.1$           & $25.9$                  & $28.5$                  & $26.4$                  \\ \midrule
        \multicolumn{3}{l}{\textbf{Extended Data Conditions}}                 &                  &                  &                         &                         &                         \\
        \multirow{5}{*}{2} & a         & 1b + More Data & $30.8$           & $30.7$           & $26.0$                  & $28.0$                  & $31.6$                  \\
                           & b         & 2a + SegAugment                      & $31.3$           & $30.9$           & $26.6$                  & $29.4$                  & $32.4$                  \\
                           & c         & 2b + synthCV                         & $\bm{31.4}$      & $\bm{31.0}$      & $26.5$                  & $29.4$                  & $32.3$                  \\
                           & d         & 2c + Knowledge Distillation          & $30.9$           & $30.7$           & $\bm{26.8}$             & $29.5$                  & $32.7$                  \\
                           & e         & 2c + higher LR                       & $31.2$           & $30.8$           & $26.4$                  & $\bm{29.8}$             & $\underline{\bm{33.4}}$ \\ \midrule
        \multicolumn{3}{l}{\textbf{Ensembles}}                                &                  &                  &                         &                         &                         \\
        \multirow{3}{*}{3} & a         & Ensemble (2d, 2e)                    & $31.4$                & $31.1$                & $26.9$                  & $29.7$                  & $32.8$                  \\
                           & b         & Ensemble (2c, 2d, 2e)                & $31.4$                 &   $31.1$               & $\underline{\bm{27.0}}$ & $\underline{\bm{29.9}}$ & $32.7$                  \\
                           & c         & Ensemble (2b, 2c, 2d, 2e)            &  $\underline{\bm{31.5}}$                &  $\underline{\bm{31.2}}$                & $\underline{\bm{27.0}}$ & $29.8$                  & $33.1$                  \\ \bottomrule
        \end{tabular}
        }
        \caption{BLEU scores for En-De MuST-C and IWSLT sets. In \textbf{bold} are the best scores by single models, and in \underline{\textbf{underlined bold}} are the best scores overall.}
        \label{tab:results}
    \end{table*}

    In Table \ref{tab:results} we provide the BLEU scores on MuST-C tst-COMMON and the IWLST test sets of tst2019 and tst2020 (TED domain), and acl2023 (ACL domain). We are using the original segmentation for MuST-C and apply SHAS with the optimal parameters (\S \ref{sec:segmentation}) of 10-26 secs for the TED domain, and 0.2-12 secs for the ACL one. We also provide the results from our submission to IWSLT '22.

    In the first part of Table \ref{tab:results}, we observe that this year's baseline (1a) improves results from last year best single model in both MuST-C and IWSLT test sets, although it only uses data from MuST-C. The reasons behind these improvements are the proper fine-tuning of learning rate and regularization parameters, as well as the choice of the speech encoder (\S \ref{sec:app_preliminary}). For the next exepriment (1b), by using the Siamese pretraining (\S \ref{subsec:ot}), instead of just using CTC for the pretraining, we obtain substantial improvements in MuST-C v2, tst2020, and acl2023, indicating the efficacy of our pretraining method when applied on top of foundation models.

    Adding more data in all parts of the training (2a), including the MT fine-tuning, Siamese pre-training and ST fine-tuning, did not bring any meaningful improvements to MuST-C and IWSLT.tst2019/20, but it dramatically improved the results on the acl2023 development set. We hypothesize that the CommonVoice and CoVoST data play an important role due to the large representation of foreign accents, similar to those in acl2023. Following, with the inclusion of SegAugment in the ST fine-tuning (2b) we observe an increase in all test sets, with larger ones in the IWSLT test sets, since SegAugment data have the same segmentation. Then, also using synthetic data from CommonVoice (2c) has minor improvements in MuST-C and a slight decrease in IWSLT. Despite that, we included synthetic data in subsequent experiments, since they were running in parallel. Applying Knowledge Distillation with the fine-tuned mBART50 as a teacher (2d), brings moderate gains of 0.1-0.4 BLEU in the IWSLT sets, and finally an increase in the learning rate (2e) from 5e-5 to 7.5e-5 provide a model that scored the best in tst2020 and acl2023.
    
    Ensembling multiple models provided small increases in all sets. We believe that there is very little variation in our best models (2b-2e), since they are initialized from the same Siamese pre-training (2b), thus resulting in ineffective ensembles. In general, and in terms of single models, we improve our results from last year by 1.6 BLEU in tst2019 and 2.1 BLEU in tst2020, while the difference is larger in terms of single models.

\section{Conclusions}

    We described the submission of the UPC Machine Translation group for the IWSLT 2023 Offline ST task. Our system leverages ASR and MT foundation models and a Siamese pretraining step to maximize the transfer learning from MT. We show that Siamese pretraining can bring significant improvements to our ST models, while fine-tuning with KD can also be helpful. We furthermore show that synthetic data are crucial at improving performance in the IWSLT test sets. In future work, we plan to investigate the zero-shot capabilities of optimal transport in the context of foundation models.

\section{Submission Results}

In Tables \ref{tab:ted2023}, \ref{tab:acl2023} and \ref{tab:sub2023}, we present the official submission results for IWSLT 2023 with our best system, which is the Ensemble 3c of Table \ref{tab:results}. Systems are evaluated on the three test sets (TED, ACL, Sub) with three metrics; BLEU \cite{bleu}, chrF \cite{chrf2}, and COMET \cite{comet}. The TED test set also has two available references.

\begin{table}[h]
\centering
\resizebox{\columnwidth}{!}{
\begin{tabular}{@{}lccccccc@{}}
\toprule
\textbf{Metric}    & \multicolumn{3}{c}{\textbf{BLEU}} & \multicolumn{2}{c}{\textbf{chrF}} & \multicolumn{2}{c}{\textbf{COMET}} \\
\textbf{Reference} & 1         & 2         & both      & 1               & 2               & 1                & 2               \\ \midrule
\textbf{System 3c} & 25.5      & 29.8      & 36.6      & 0.56            & 0.58            & 0.7985           & 0.8098          \\ \bottomrule
\end{tabular}
}
\caption{Official Results for the TED test set 2023.}
\label{tab:ted2023}
\end{table}

\begin{table}[h]
\centering
\resizebox{0.7\columnwidth}{!}{
\begin{tabular}{@{}lccc@{}}
\toprule
\textbf{Metric}    & \textbf{BLEU} & \textbf{chrF} & \textbf{COMET} \\ \midrule
\textbf{System 3c} & 32.1          & 0.6           & 0.7473         \\ \bottomrule
\end{tabular}
}
\caption{Official Results for the ACL test set 2023.}
\label{tab:acl2023}
\end{table}

\begin{table}[h]
\centering
\resizebox{0.7\columnwidth}{!}{
\begin{tabular}{@{}lccc@{}}
\toprule
\textbf{Metric}    & \textbf{BLEU} & \textbf{chrF} & \textbf{COMET} \\ \midrule
\textbf{System 3c} & 15.6          & 0.47           & 0.3746         \\ \bottomrule
\end{tabular}
}
\caption{Official Results for the Sub test set 2023.}
\label{tab:sub2023}
\end{table}

% \newpage

\section*{Acknowledgements}

    The work done by Ioannis Tsiamas and Gerard I. Gállego was supported by the ADAVOICE project, PID2019-107579RB-I00 / AEI / 10.13039/501100011033

\bibliography{anthology,custom}
\bibliographystyle{acl_natbib}

\appendix

\section{Appendix}
\label{sec:appendix}

\subsection{Implementation Details}
\label{subsec:impl_details}

    This section presents the implementation details of our proposed model architecture.
    
    As an ASR model, we are using wav2vec 2.0\footnote{\url{https://dl.fbaipublicfiles.com/fairseq/wav2vec/wav2vec2_vox_960h_new.pt}} which is composed of a 7-layer convolutional feature extractor and 24-layer Transformer encoder. It is pretrained with 60k hours of non-transcribed speech from Libri-Light \citep{librilight}, and fine-tuned for ASR with 960 hours of labeled data from Librispeech \citep{librispeech}. The wav2vec 2.0 version we use was also fine-tuned with pseudo-labels \citep{wav2vec-selftraining}.

    As an MT model, we are using mBART50 \cite{mbart50}, which is already fine-tuned on En-Xx multilingual machine translation\footnote{\url{https://dl.fbaipublicfiles.com/fairseq/models/mbart50/mbart50.ft.1n.tar.gz}}. We further pretrain it for two reasons. Firstly, we are only interested in the En-De direction, and thus we would like a more specialized model on that direction. Secondly, due to the 2nd step of encoder matching, we would like the text encoder to have a very good representation of our data. For MT fine-tuning, we use the original parameters of mBART50 \cite{mbart50}, and the datasets listed in Table \ref{tbl:mt_data}.
    
    The acoustic encoder has 24 Transformer layers, while the semantic encoder and the decoder have 12 layers each. All layers have an embedding dimensionality of 1024, a feed-forward dimensionality of 4098, GELU activations \cite{gelu}, 16 attention heads, and pre-layer normalization \cite{prelayernorm}. The vocabulary for the CTC has a size of 32 characters, while the one for the ST model has a size of 250,000. 

    The model takes waveforms with a 16kHz sampling rate as input, which are normalized to zero mean and unit variance. The models are trained using the data presented in Table \ref{tbl:train_data}, with maximum source length of 400,000 and target length of 1024 tokens. Gradient accumulation and data parallelism are employed to achieve an effective batch size of approximately 32 million tokens.

    For the Siamese pre-training we use Adam \cite{adam} with a base learning rate of $2 \cdot 10^{-4}$, a warm-up of 1,000 steps and an inverse square root scheduler. We follow a reduced regularization approach, as compared to the original configuration of wav2vec 2.0 and mBART50, which we found to work the best in our preliminary experiments. Thus, we use 0.1 activation dropout in the acoustic encoder, as well as time masking with probability of 0.2 and channel masking with probability of 0.1. For the context encoder, we use 0.1 dropout and 0.1 attention dropout. All other dropouts are inactive. All the weights in the loss function were set to 1.0 (Eq. \ref{eq:ot_loss}). We train until the $\mathcal{L}^{OT_2}$ term of the loss does not improve for 5,000 steps, and then average the 10 best checkpoints according to the same loss term.

    For ST fine-tuning, we use Adam with a base learning rate of $5 \cdot 10^{-5}$, fixed for the 20$\%$ of the training before decaying to $5 \cdot 10^{-7}$ for the rest. In the semantic encoder, we apply a dropout of 0.1 and an attention dropout of 0.1, while for the decoder we use a dropout of 0.3 and an attention dropout of 0.1. Neither dropout nor masking is applied in the frozen acoustic encoder. The loss is the cross-entropy with label smoothing of 0.2.
    
    For the experiments incorporating Knowledge Distillation (KD) during ST fine-tuning, the loss is calculated as a weighted sum of the standard cross-entropy (no label smoothing) and the KL divergence between the teacher and student distributions, controlled by a hyperparameter $\lambda$, set to 0.5. The teacher distribution for each step is obtained offline using the fine-tuned mBART50, where we keep the top-8 indices, and both the teacher and student distributions are additionally modified with temperature $T=1.3$ \cite{kd2020}.

    After ST fine-tuning, we pick the 10 best checkpoints according to the BLEU \citep{bleu} computed with sacreBLEU \cite{sacrebleu} on the development set of MuST-C and average them. For generation, we use a beam search of 5. All models are implemented in \textsc{fairseq} \cite{fairseq}, and experiments were run on a cluster of 8 NVIDIA GeForce RTX 3090. Our code is available at a public repository\footnote{\url{https://github.com/mt-upc/iwslt-2023}}.

\subsection{MT fine-tuning}
\label{sec:app_mt}

For the MT fine-tuning, we use the parallel text of the ST datasets, as well as Europarl v10 En-De \cite{europarlmt} (Table \ref{tbl:mt_data}). We perform text normalization and remove pairs with extremely short text segments (fewer than 4 characters) or extreme source-to-target length ratio (less than 0.5 or larger than 2).

    \begin{table}[h]
        \centering
        \resizebox{0.7\columnwidth}{!}{
        \begin{tabular}{lcc}
        \toprule
                                            & \textbf{Original}     & \textbf{Filtered}   \\
        \midrule
        \textbf{ST datasets}                &                       &               \\
        MuST-C v3                           & $270$                 & $235$        \\
        Europarl-ST                         & $33$                  & $26$         \\
        CoVoST 2                            & $231$                 & $203$        \\
        \midrule
        \textbf{MT datasets}                &                       &               \\
        Europarl v10                        & $1,829$               & $1,566$       \\
        \midrule
        \textbf{Total}                      & $2,363$               & $2,030$       \\
        \bottomrule
        \end{tabular}
        }
        \caption{Filtered training data (thousands of sentences) for MT fine-tuning stage.}
        \label{tbl:mt_data}
    \end{table}

    \begin{table}[h]
        \centering
        \resizebox{0.99\columnwidth}{!}{
        \begin{tabular}{lcccc}
        \toprule
                                       & \multicolumn{2}{c}{\textbf{\textbf{MuST-C}}} & \multirow{2}{*}{\textbf{\textbf{Europarl-ST}}} & \multirow{2}{*}{\textbf{\textbf{CoVoST2}}} \\
                                       & \textbf{v2}            & \textbf{v3}            &                                                &                                            \\ \midrule
        \textbf{Off-the-shelf}         &                        &                        &                                                &                                            \\
        mBART50                        & $31.4$                 & $30.9$                 & $35.0$                                         & $33.6$                                     \\ \midrule
        \textbf{Fine-tuned}            &                        &                        &                                                &                                            \\
        MuST-C v2                      & $35.3$                 & $34.4$                 & $34.6$                                         & $35.3$                                     \\
        All (\S \ref{subsec:datasets}) & $34.9$                 & $34.2$                 & $40.3$                                         & $39.9$                                     \\ \bottomrule
        \end{tabular}
        }
        \caption{BLEU scores on MT test sets.}
        \label{tbl:mt_results}
    \end{table}

\subsection{Preliminary experiments}
\label{sec:app_preliminary}

    Before starting the primary experiments for the IWSLT evaluation campaign, we conducted an array of preliminary tests, building on top of previous years' submissions \cite{upc2021,upc2022}. These explorations were intended to examine the impact of system configuration variations on the performance metrics on the MuST-C v2 dev set, such as BLEU \cite{bleu}, chrF2 \cite{chrf2}, and COMET \cite{comet}. To ensure the robustness of our findings, we estimated statistical significance using the bootstrap resampling method \cite{bootstrap-resampling}.

    In our initial experiment, we examined the impact of various fine-tuning strategies used in our last years' participations, specifically \textit{LNA} \cite{lna} and \textit{LNA-Adapters} \cite{upc2022}, in comparison to full fine-tuning. The goal was to verify whether these approaches inadvertently hurt the system's performance. As demonstrated in Table \ref{tab:expl_lna}, these strategies indeed had a detrimental effect, leading to reductions of 1.9 BLEU points when applied to both the encoder and the decoder. Consequently, we opted to adopt a conventional full fine-tuning strategy for subsequent experiments.
    
    Following this, we conducted a comparative analysis of various speech encoders, including different variations of \textit{wav2vec 2.0} \cite{wav2vec2.0,wav2vec-selftraining,wav2vec-robust,xlsr}, \textit{HuBERT} \cite{hubert}, and \textit{SpeechLM} \cite{speechlm} (Table \ref{tab:expl_spenc}). Our baseline was the wav2vec 2.0 fine-tuned with pseudo-labels \cite{wav2vec-selftraining}, and intriguingly, most encoders exhibited a comparable level of performance. A marginal decrease was observed with the wav2vec 2.0 pretrained on a large pool of datasets (LV-60 + CV + SWBD + FSH) \cite{wav2vec-robust}, and the multilingual version of wav2vec 2.0, XLSR \cite{xlsr}. The SpeechLM results were noticeably below expectations, leading us to suspect a bug in our implementation.
    
    Upon noting that the hyperparameters were optimized for a specific speech encoder, we hypothesized that a reduction in the learning rate might boost HuBERT's performance. However, as demonstrated in Table \ref{tab:expl_lr}, the performance was adversely affected, prompting us to retain the original wav2vec 2.0 as the primary speech encoder due to the lack of substantial improvements offered by other alternatives.
    
    \begin{table}
        \centering
        \resizebox{0.45\textwidth}{!}{
        \begin{tabular}{ccccc}
        \toprule
        \textbf{Encoder}    & \textbf{Decoder}  & \textbf{BLEU}     & \textbf{chrF2}    & \textbf{COMET}        \\
        \midrule
        -                   & -                 & $29.0$            & $54.7$            & $0.8001$              \\
        \midrule
        LNA                 & -                 & $\;\;28.0\,^*$    & $\;\;54.1\,^*$    & $\;\;0.7949\,^*$      \\
        -                   & LNA               & $\;\;27.9\,^*$    & $\;\;54.0\,^*$    & $\;\;0.7882\,^*$      \\
        LNA                 & LNA               & $\;\;27.1\,^*$    & $\;\;53.2\,^*$    & $\;\;0.7800\,^*$      \\
        \midrule
        LNA-Adapt           & -                 & $\;\;28.2\,^*$    & $\;\;54.3\,^*$    & $\;\;0.7960\,^*$      \\
        -                   & LNA-Adapt         & $\;\;27.6\,^*$    & $\;\;53.6\,^*$    & $\;\;0.7889\,^*$      \\
        LNA-Adapt           & LNA-Adapt         & $\;\;27.1\,^*$    & $\;\;53.5\,^*$    & $\;\;0.7847\,^*$      \\
        \bottomrule
        \end{tabular}
        }
        \caption{Performance comparison of fine-tuning strategies w.r.t. to full fine-tuning, evaluated on the MuST-C v2 dev set (en-de). \textit{LNA} and \textit{LNA-Adapters} represent the strategies proposed by \cite{lna} and \cite{upc2022} respectively. $*$ indicates significance w.r.t. baseline (full fine-tuning).}
        \label{tab:expl_lna}
    \end{table}

    \begin{table*}[b]
        \centering
        \resizebox{0.75\textwidth}{!}{
        \begin{tabular}{lcccc}
        \toprule
        \textbf{System} & \textbf{ASR FT}   & \textbf{BLEU} & \textbf{chrF2}  & \textbf{COMET}  \\
        \midrule
        Wav2Vec 2.0 Large (LV-60) + Self Training                   & \cmark    & $30.2$            & $56.1$            & $0.8087$      \\
        \midrule
        Wav2Vec 2.0 Large (LV-60)                                   & \cmark    & $30.1$            & $55.9$            & $0.8098$      \\
        Wav2Vec 2.0 Large (LV-60)                                   & \xmark    & $30.3$            & $55.9$            & $-$           \\
        Wav2Vec 2.0 Large (LV-60 + CV + SWBD + FSH)                 & \cmark    & $\;\;29.7\,^*$    & $\;\;55.7\,^*$    & $0.8083$      \\
        Wav2Vec 2.0 Large (LV-60 + CV + SWBD + FSH)                 & \xmark    & $30.0$            & $55.9$            & $-$           \\
        Wav2Vec 2.0 Large conformer - rope (LV-60)$\,^{\dagger}$    & \cmark    & $29.8$            & $\;\;55.4\,^*$    & $-$           \\
        XLSR-53                                                     & \xmark    & $\;\;28.9\,^*$    & $\;\;55.0\,^*$    & $-$           \\
        \midrule
        HuBERT Large                                                & \cmark    & $30.3$            & $56.1$            & $0.8099$      \\
        HuBERT Large                                                & \xmark    & $30.3$            & $56.2$            & $0.8110$      \\
        \midrule
        SpeechLM-P Large$\,^{\ddagger}$                              & \xmark    & $\;\,23.6\,^*$    & $\;\,50.2\,^*$    & $-$           \\
        \bottomrule
        \end{tabular}
        }
        \caption{Speech encoders exploration with MuST-C v2 dev set (en-de). $*$ indicates significance w.r.t. baseline (1st row). $\dagger$ uses \textit{LNA-Adapters} \cite{upc2022}. $\ddagger$ indicates a possible bug in our implementation.}
        \label{tab:expl_spenc}
    \end{table*}

    \begin{table*}
        \centering
        \resizebox{0.7\textwidth}{!}{
        \begin{tabular}{cccccc}
        \toprule
        \textbf{Encoder Reg.}   & \textbf{Decoder Reg.} & \textbf{WavAugm.}     & \textbf{BLEU}     & \textbf{chrF2}    & \textbf{COMET}    \\
        \midrule
        base                    & base                  & \cmark                & $30.2$            & $56.1$            & $0.8087$          \\
        \midrule
        base                    & original              & \cmark                & $30.5$            & $\;\;56.4\,^*$    & $\;\;0.8149\,^*$  \\
        base                    & original              & \xmark                & $30.7$            & $\;\;56.4\,^*$    & $\;\;0.8127\,^*$          \\
        base                    & reduced               & \cmark                & $30.1$            & $55.9$            & $0.8078$          \\
        \midrule
        original                & base                  & \cmark                & $29.8$            & $55.8$            & $0.8100$          \\
        reduced                 & base                  & \cmark                & $30.1$            & $55.9$            & $0.8108$          \\
        \midrule
        original                & original              & \cmark                & $30.4$            & $56.2$            & $\;\;0.8138\,^*$  \\
        reduced                 & reduced               & \cmark                & $30.1$            & $56.0$            & $\;\;0.8122\,^*$  \\
        \bottomrule
        \end{tabular}
        }
        \caption{Variations of the regularization and data augmentation strategies, with MuST-C v2 dev set (en-de). $*$ indicates significance w.r.t. baseline (1st row).}
        \label{tab:expl_reg}
    \end{table*}
    
    \begin{table}
        \centering
        \resizebox{0.45\textwidth}{!}{
        \begin{tabular}{cccc}
        \toprule
        \textbf{Learning Rate}  & \textbf{BLEU}     & \textbf{chrF2}    & \textbf{COMET}    \\
        \midrule
        $5\cdot10^{-4}$         & $30.3$            & $56.1$            & $0.8099$          \\
        \midrule
        $2\cdot10^{-4}$         & $30.3$            & $56.0$            & $0.8069$          \\
        $1\cdot10^{-4}$         & $30.2$            & $55.9$            & $0.8085$          \\
        $5\cdot10^{-5}$         & $\;\;29.5\,^*$    & $\;\;55.3\,^*$    & $0.8047$          \\
        \bottomrule
        \end{tabular}
        }
        \caption{Learning rate search for HuBERT encoder, with MuST-C v2 dev set (en-de). $*$ indicates significance w.r.t. baseline (1st row).}
        \label{tab:expl_lr}
    \end{table}
    
    \begin{table}
        \centering
        \resizebox{0.45\textwidth}{!}{
        \begin{tabular}{lccc}
        \toprule
        \textbf{Training Data}  & \textbf{BLEU}     & \textbf{chrF2}    & \textbf{COMET}    \\
        \midrule
        MuST-C v2               & $30.7$            & $56.4$            & $0.8127$          \\
        \midrule
        MuST-C v3               & $30.5$            & $56.6$            & $0.8118$          \\
        \bottomrule
        \end{tabular}
        }
        \caption{Performance of the systems trained with different versions of MuST-C, evaluated with MuST-C v2 dev set (en-de). No significant improvements found.}
        \label{tab:expl_mustc3}
    \end{table}

    Our focus then shifted towards examining the influence of varying regularization and data augmentation strategies on system performance (Table \ref{tab:expl_reg}). We explored a range, from our traditionally used setup (\textit{base}), to the one employed in the \textit{original} foundation model fine-tuning, and a \textit{reduced} version. Implementing the \textit{original} regularization within the speech encoder, as opposed to the \textit{base} variant, significantly boosted performance, leading us to select this configuration. We also explored the effectiveness of WavAugment \cite{wavaugment}, ultimately finding that, despite its training speed slowdown, it did not enhance the results. Consequently, we opted to stop using it.
    
    Lastly, we evaluated the potential benefits of employing the new MuST-C v3 training data on system performance (Table \ref{tab:expl_mustc3}). Unexpectedly, no significant improvements were observed upon transitioning from MuST-C v2 to v3. Despite this, we decided to utilize v3, since it's specifically prepared for the IWSLT evaluation campaign.

    These preliminary investigations have not only provided a more profound understanding of the role of each system's component and setting, but also have yielded us with a better starting point for the subsequent experiments of our work.

\end{document}